\PassOptionsToPackage{square,comma,numbers,sort&compress}{natbib}
\documentclass{article}

\usepackage[final]{neurips_2024}

\usepackage[utf8]{inputenc} 
\usepackage[T1]{fontenc}    
\usepackage{xcolor}         
\definecolor{cvprblue}{rgb}{0.21,0.49,0.74}
\usepackage[pagebackref,breaklinks,colorlinks,citecolor=cvprblue]{hyperref}
\usepackage{url}            
\usepackage{booktabs}       
\usepackage{amsfonts}       
\usepackage{nicefrac}       
\usepackage{microtype}      
\usepackage{xcolor}         

\usepackage{graphicx}
\usepackage{caption}
\usepackage{subcaption}
\usepackage{xspace}
\usepackage{float}
\usepackage{wrapfig}
\usepackage{diagbox}
\usepackage{multirow}

\title{Unsupervised Modality Adaptation with Text-to-Image Diffusion Models for Semantic Segmentation}

\author{
    Ruihao Xia$^{1,2}$\thanks{Work was done during interning at vivo.}\,\,,  Yu Liang$^2$\,\,, Peng-Tao Jiang$^2$\,\,, Hao Zhang$^2$ \\ 
    \textbf{Bo Li}$^2$\thanks{Corresponding author.}\,\,, \textbf{Yang Tang}$^{1,3}$\footnotemark[2]\,\,, \textbf{Pan Zhou}$^4$ \\
     $^1$East China University of Science and Technology, $^2$vivo Mobile Communication Co., Ltd\\
     $^3$Peng Cheng Laboratory,
     $^4$Singapore Management University
}

\begin{document}

	\maketitle

	\begin{abstract}

		Despite their success, unsupervised domain adaptation methods for semantic segmentation primarily focus on adaptation between image domains and do not utilize other abundant visual modalities like depth, infrared and event. This limitation hinders their performance and restricts their application in real-world multimodal scenarios. To address this issue, we propose Modality Adaptation with text-to-image Diffusion Models (MADM) for semantic segmentation task which utilizes text-to-image diffusion models pre-trained on extensive image-text pairs to enhance the model's cross-modality capabilities.   
		Specifically, MADM comprises two key complementary components to tackle major challenges. First, due to the large modality gap, using one modal data to generate pseudo labels for another modality suffers from a significant drop in accuracy. To address this, MADM designs diffusion-based pseudo-label generation which adds latent noise to stabilize pseudo-labels and enhance label accuracy. Second, to overcome the limitations of latent low-resolution features in diffusion models, MADM introduces the label palette and latent regression which converts one-hot encoded labels into the RGB form by palette and regresses them in the latent space, thus ensuring the pre-trained decoder for up-sampling to obtain fine-grained features. Extensive experimental results demonstrate that MADM achieves state-of-the-art adaptation performance across various modality tasks, including images to \textit{depth}, \textit{infrared}, and \textit{event} modalities. We open-source our code and models at \url{https://github.com/XiaRho/MADM}.
	\end{abstract}

	\section{Introduction}
	\label{Introduction}

        \begin{figure}[t]
        \centering
        \centerline{\includegraphics[scale=0.6]{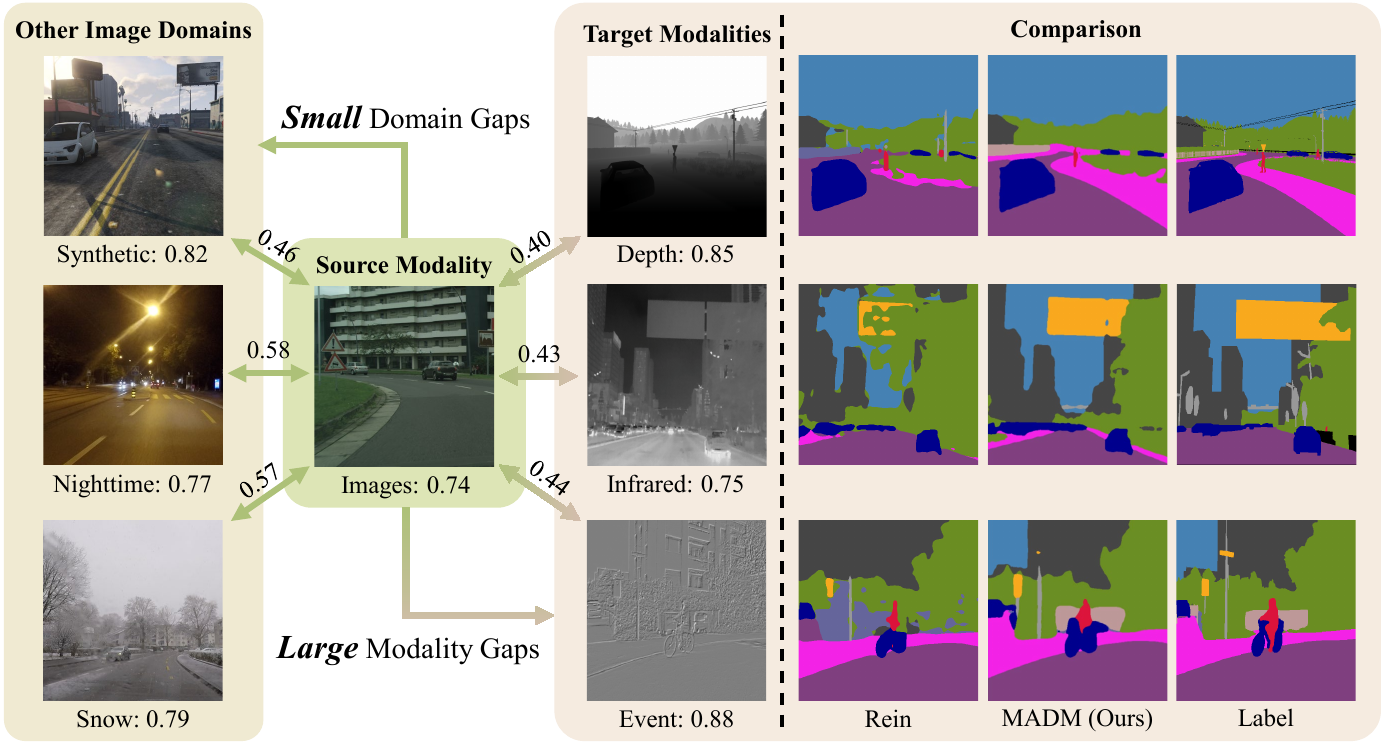}}
        \caption{
            (1) On the left, we leverage the multi-modality model ImageBind~\cite{ImageBind} to quantify the similarity of images and modalities across datasets, i.e., GTA5-Synthetic~\cite{GTA5}, Dark Zurich-Nighttime~\cite{DarkZurich}, ACDC-Snow~\cite{ACDC}, DELIVER-Depth~\cite{DELIVER}, FMB-Infrared~\cite{FMB}, and DSEC-Event~\cite{DSEC}. 
            Specifically, we randomly select 500 samples from each dataset, and compute the average cosine similarity of the output vectors within the dataset (right side of the text) and between the datasets (on the arrows). 
            (2) On the right, we compare the quantitative results with the state-of-the-art (SoTA) method Rein~\cite{Rein} on three different modalities.
        }
        \label{fig:intro}
        \end{figure}
    
	Unsupervised Domain Adaptation for Semantic Segmentation (UDASS) involves a source domain with image-label pairs and a target domain with only unlabeled samples~\cite{SS_st_1,SS_depth,SS_pixel}, and has achieved promising segmentation results in the image modality.   Currently, most existing UDASS methods are restricted to transferring knowledge between similar image domains, such as from virtual scene~\cite{GTA5,Synthia} to real scene~\cite{Cityscapes}, or from daytime scene~\cite{Cityscapes} to nighttime scene~\cite{DarkZurich,ACDC}.  
	However, these approaches do not account for the wide range of visual modalities present in real-world scenarios, such as depth, infrared, and event modalities, which are valuable in nighttime perception~\cite{DepthNight,InfraredNight,CMDA} but often lack sufficient and high-quality labels for supervising segmentation training.
	Hence, in this paper, we are particularly interested in extending UDASS to Unsupervised Modality Adaptation for Semantic Segmentation (UMASS) across different visual modalities, i.e., the adaptation of a model from a labeled source image modality to an unlabeled target modality.
	
	Differences across images arise from the objects, lighting, camera parameters, etc, while there are fundamental disparities in imaging principles across modalities that lead to greater variability.
 	This is illustrated by  Figure~\ref{fig:intro} which shows that the similarity among various image domains tends to be higher than between different modalities. These significant modality discrepancies poses significant challenges to existing UDASS methods~\cite{DAFormer,MIC} on  multimodal  segmentation due to their limited pre-trained knowledge. 
	Specifically, the backbone~\cite{SegFormer} used in current SoTA UDASS methods is pre-trained on the ImageNet-1K dataset~\cite{ImageNet} which contains one million images categorized into 1,000 distinct classes, providing the network with a foundational level of semantic understanding. While this backbone achieves promising results in UDASS, its insufficient pre-trained knowledge limits generalization to other visual modalities. 
	
	To address this issue, inspired by Text-to-Image Diffusion Models (TIDMs)~\cite{LDM}, which are trained with internet-scale image-text pairs~\cite{LAION-5B}, we recognize that extensive pre-training data unifies samples with different distributions but similar semantic properties through texts, significantly enhancing the model's semantic understanding and generalization.
 	Although TIDMs are not trained on other visual modalities, their large-scale samples and unification through texts enable them to adapt to a broader distribution of domains. Also, the extensive pretraining provides TIDMs with a robust understanding of high-level visual concepts, enabling their application to various domains, such as semantic matching~\cite{SD4Match}, depth estimation~\cite{Marigold}, and 3D awareness~\cite{3DAwareness}. This strong prior motivates us to utilize TIDMs as a robust backbone for solving UMA. 
	Therefore, we present MADM: Modality Adaptation with text-to-image Diffusion Models, which takes full advantage of the generalization of pre-trained diffusion models and facilitates robust adaptation for accurate semantic segmentation in other visual modalities. 
 	Specifically,  TIDM is used to extract robust features for segmentation and is trained in a self-training manner~\cite{DAFormer} which takes unlabeled target samples as input and generates pseudo-labels for training TIDM itself.
	Building upon TIDMs~\cite{LDM} and self-training~\cite{DAFormer}, our MADM incorporates two innovative components: Diffusion-based Pseudo-Label Generation (DPLG) and Label Palette and Latent Regression (LPLR), which address the challenges of unstable pseudo-labeling and lack of fine-grained features extraction, respectively.

	First, significant modality discrepancies hinder robust and high-quality pseudo-label generation, which is crucial for further training and enhancing the model. Directly using these unstable labels to train the model can lead to serious biases in the target modality. Thus, we propose DPLG which adds latent noise to target samples before generating pseudo-labels where the noise level gradually decreases as training stabilizes.  
 	These pseudo-labels then supervise the noise-free target modality predictions. The mechanism behind DPLG leverages the denoising property of diffusion models, making the target modality more consistent with pre-trained inputs, thus improving accuracy. Unlike previous supervised diffusion-based semantic segmentation methods~\cite{VPD,ODISE,PTDiffSeg} which adopt a single-step forward and remove the diffusion process, we find that a proper diffusion process can stabilize the generation of pseudo-labels and adapt more successfully to the target modality.
   
	Second, TIDMs~\cite{LDM} encode images into the latent space using a pre-trained Variational AutoEncoder (VAE) for diffusion and denoising, and then up-sample the latent output to the original resolution using the VAE decoder. When adopting TIDMs as a backbone, the resolution of the features is too low, resulting in a loss of details. To address this, we propose LPLR to  convert pixel-level classification into regression in RGB form,  utilizing the up-sampling capability of the pre-trained VAE decoder in a recycling manner. Specifically, we use a palette to convert one-hot encoded labels into RGB form and encode these RGB labels into latent representations. Then, the model is  trained with a regression loss to fit the latent labels, obtaining high-resolution fine-grained features via the VAE decoder. Different from previous diffusion-based semantic segmentation methods~\cite{VPD,ODISE,PTDiffSeg} that directly extract multi-scale features from the denoising UNet network, we take inspiration from depth estimation with TIDMs~\cite{Marigold} and propose LPLR to extract high-resolution features. Our contributions are summarized as follows:
	
	\begin{itemize}
	\item We propose MADM, extending traditional UDASS to UMASS with a pre-trained TIDM backbone for generalizing across various visual modalities.
	\item	We design  Diffusion-based Pseudo-Label Generation (DPLG) to provide more robust pseudo-labels by adding annealed latent noise to target samples for stable modality adaptation.
	\item	We introduce  Label Palette and Latent Regression (LPLR) to convert semantic segmentation into regression for learning  details, thereby repeatedly utilizing pre-trained VAE decoders for high-resolution feature extraction.
	\item	We demonstrate the effectiveness of our MADM through extensive experiments on three different modalities: Image~\cite{Cityscapes} $\to$ Depth~\cite{DELIVER}, Infrared~\cite{FMB}, and Event~\cite{DSEC}.
 	\end{itemize}
	
	\section{Related Works}
	
	\subsection{Unsupervised Domain Adaptation Semantic Segmentation}
	
	UDASS can be broadly categorized into two primary methods: adversarial and self-training methods. 
	Adversarial approaches aim to align the distributions of the source and target domains at the level of images~\cite{SS_adv_input1,SS_adv_input2} or features~\cite{SS_adv_feat1,SS_adv_feat2} or outputs~\cite{SS_adv_output1,SS_adv_output2}, thereby facilitating the transfer of knowledge. 
	On the other hand, self-training methods~\cite{SS_st_1,SS_st_2} operate on the paradigm of pseudo-labeling, where the model's predictions on the target domain act as ground truth during training, iteratively refining the segmentation. 
	Recently, the field has witnessed the development from CNN-based methods~\cite{SS_adv_input2} to Transformer-based methods~\cite{DAFormer,MIC,HRDA}, leveraging the self-attention mechanism to capture long-range dependencies and enhance cross-domain feature representation. 
	
	However, most of these methods have predominantly focused on adaptation between different image domains, such as from synthetic~\cite{GTA5,Synthia} to real-world images or across varying environmental conditions~\cite{ACDC}.
	They will fail when adapting to other visual modalities, such as depth, infrared, or event, which have distinct data distribution. 
	Thus, we propose MADM to address the limitation of UDASS for adapting to other unexplored visual modalities.
	Besides, different from multi/cross-modality domain adaptation~\cite{MMDA1,MMDA2} which has paired two modalities in both domains, our modalities are different in source and target.
	
	\subsection{Text-to-Image Diffusion Models}
	
	Diffusion denoising probabilistic models~\cite{DM} have set new benchmarks in the quality and controllability of generative tasks. 
	These models are distinguished by their two-phase paradigm: the diffusion process that progressively adds noise to the sample, and the denoising process that learns to denoise the corrupted sample by a network. 
	Utilizing the MSE loss between the residual noise and the prediction as a training objective, diffusion models have demonstrated greater training stability compared to generative adversarial networks~\cite{GAN} and VAEs~\cite{VAE}. 
	To achieve high-quality controllable image generation with reduced computational demands, Rombach \itshape{et al.}\upshape~\cite{LDM} proposed the latent diffusion model that leverages cross-attention layers and confines the diffusion process to a low-resolution latent space, which has emerged as a widely recognized TIDM.
	
	In recent advancements beyond the generative tasks, the exceptional semantic comprehension capabilities of TIDMs have been harnessed to enhance performance in many downstream applications. 
	Xu \itshape{et al.}\upshape~\cite{ODISE} presented a novel framework that integrates a pretrained TIDM with a discriminative model to address the challenges of open-vocabulary segmentation. 
	Similarly, Zhao \itshape{et al.}\upshape~\cite{VPD} demonstrated the versatility of TIDMs by fine-tuning it to deal with various visual perception tasks, including semantic segmentation, referring image segmentation, and depth estimation. 
	Gong \itshape{et al.}\upshape~\cite{PTDiffSeg} introduced innovative scene prompts and a prompt randomization strategy on TIDMs, achieving new milestones in domain generalization and test-time domain adaptation. 
	Their works highlight the potential of TIDMs to generalize across diverse domains and adapt to new ones with minimal additional training. 
	It's worth noting that the aforementioned methods necessitate only a single-step forward pass through the denoising UNet, significantly streamlining the inference process. 
	
	The successful and diverse applications of TIDMs inspire our exploration into the generalization of TIDMs to more challenging visual modalities. 
	However, we observe that the latent diffusion property within TIDMs leads to the lower-resolution feature extraction.
	To address this limitation, we propose the LPLR that converts semantic segmentation into latent regression to obtain fine-grained features. 

        \begin{figure}[t]
		\centering
		\centerline{\includegraphics[scale=0.51]{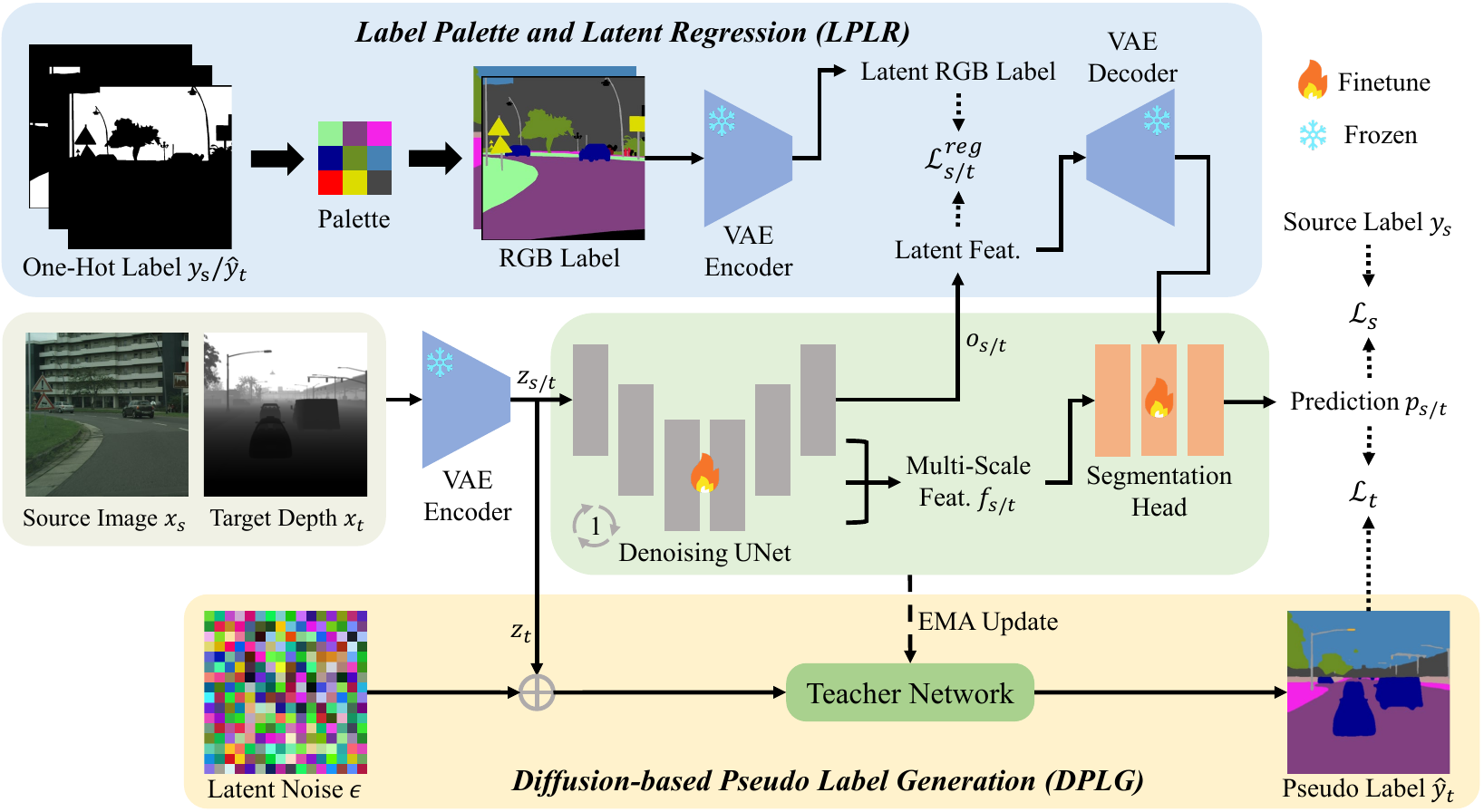}}
		\caption{
			Our framework is divided into three parts. (1) Self-Training: Supervised loss in the source modality $\mathcal{L}_{s}$ and pseudo-labeled loss $\mathcal{L}_{t}$ in the target modality are used to train the network.
			(2) Diffusion-based Pseudo-Label Generation (DPLG): In the early stage of training, we add noise on the latent representation $z_t$ to stabilize the pseudo-label generation.
			(3) Label Palette and Latent Regression (LPLR): The one-hot encoded labels $y_s/\hat{y}_t$ are converted to RGB form by palette and then encoded to the latent space to supervise the UNet output $o_{s/t}$.
		}
		\label{fig:framework}
	\end{figure}
 
	\section{Method}
	
	\subsection{Overview} 
	
	In UMASS, given the labeled source RGB modality $\left\{(x_s, y_s)\right\}$ and the unlabeled target modality $\left\{(x_t)\right\}$, our objective is to train a network which accepts $x_t$ as input and outputs the corresponding semantic segmentation results $p_t$. 
	As shown in Sec.~\ref{Introduction}, the primary challenge in this task stems from the significant disparities between the two modalities. 
	To address this challenge, we propose to leverage the TIDM~\cite{LDM} as our backbone which is pre-trained on a vast array of image-text pairs~\cite{LAION-5B} to enhance its generalization and can robustly extract features across modalities. 
	Next, to overcome the inaccurate pseudo-labels due to large modality gaps, we propose Diffusion-based Pseudo-Label Generation (DPLG), which stabilizes pseudo-label generation by injecting noise to the target modality.
	Moreover, TIDMs can only extract low-resolution features within the latent space, leading to a loss of semantic detail. 
	To address this, we propose the Label Palette and Latent Regression (LPLR), which transforms pixel-wise classification into the regression, thereby allowing us to harness the fine-grained features upsampled by the pre-trained VAE Decoder. 
	Our framework is illustrated in Figure~\ref{fig:framework}. Next, we will introduce the our proposed DPLG and LPLR in detail.

	\subsection{Diffusion-based Pseudo-Label Generation}

	As shown in Figure~\ref{fig:framework}, in MADM, we employ the TIDM to perform a single-step diffusion to extract multi-scale features from the intermediate output of the denoising UNet, following the approach in~\citep{ODISE,VPD}. First, samples from source and target modalities $x_s, x_t$ are encoded to the latent representation $z_{s/t} = \mathcal{E}(x_{s/t})$ by the pretrained VAE encoder $\mathcal{E}$. Without any additional noise, we feed them to the denoising UNet and obtain multi-scale features $f_{s/t}=\mathrm{UNet}(z_{s/t}, c)$, where $c$ is a learnable conditioned embedding instead of a textual description. Then, these features are subsequently fed into a segmentation head to generate the semantic segmentation prediction $p_{s/t}=\mathrm{Seg}(f_{s/t})$.

	Our training method is anchored on the self-training DAFormer~\cite{DAFormer} prevalent in UDASS. The training objective is a composite of supervised loss from the source modality and pseudo-labeled loss from the target modality. For the labeled source modality, we use a cross-entropy loss between the prediction $p_s$ and the ground truth labels $y_s$:
	\[ \mathcal{L}_{s} = \mathcal{L}_{CE}(p_s, y_s). \]
	
	For the unlabeled target modality, we adopt a student-teacher paradigm in self-training~\cite{DAFormer}. Here, the existing network acts as the student, and through the Exponential Moving Average (EMA), we derive a teacher network~\cite{EMA}. The teacher network generates pseudo-labels $\hat{y}_t$, which then supervise the student network's output on $A(x_t)$, where $A(\cdot)$ represents the strong data augmentation~\cite{DACS}. 

	However, the data distribution varies greatly between modalities. The teacher network is unable to provide accurate pseudo-labels for self-training, resulting in an unstable modality adaptation to the target modality. We observe that more robust pseudo-labels can be generated by injecting appropriate noise in the latent space and therefore propose the Diffusion-based Pseudo-Label Generation (DPLG). The proposed DPLG exploits the denoising property (perception of noise) of diffusion models to improve the robustness of semantic understanding on target samples.

	Specifically, given a target sample $x_t$, we first encode it into the latent representation $z_t=\mathcal{E}(x_t)$ and then apply a diffusion process that adds noise $\epsilon$ to $z_t$ to obtain a noisy latent representation:
	\[ z_t' = \sqrt{\bar{\alpha}_k}z_t + (1-\sqrt{\bar{\alpha}_k})\epsilon,~~\epsilon \sim \mathcal{N}(0, I),~~k=\beta \cdot \mathrm{max}(0, 1 - i/ \gamma ). \]

        \begin{wrapfigure}{r}{0.49\textwidth}
		\vspace{-13.5pt}
		\centering
		\includegraphics[width=0.47\textwidth]{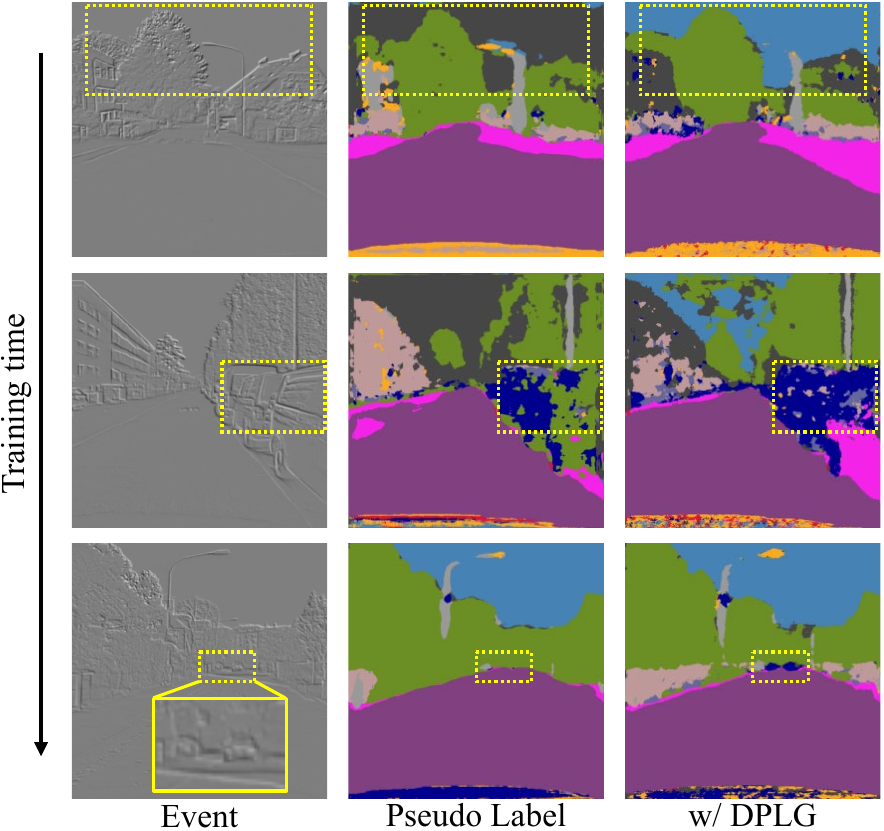}
		\captionsetup{font={scriptsize}} 
		\vspace{-6pt}
		\caption{
			We visualize the pseudo-labels for event modality at the iteration of 1250, 1750, and 2250. The introduction of DPLG effectively improves the quality of pseudo-labels.
		}
		\label{fig:DPLG}
		\vspace{-20pt}
	\end{wrapfigure}
 
	Here, $\bar{\alpha}_k$ is a predetermined schedule that controls the amount of noise added at each step~\cite{DM}. $k$ is the diffusion step that controls the proportion of noise based on the initial diffusion step $\beta$ and noise addition period $\gamma$, and $i$ is the current iteration count.
	
	This noisy representation $z_t'$ is then used to generate pseudo-labels $\hat{y}_t = \mathrm{Seg}(\mathrm{UNet}(z_t', c))$. In the pre-training of TIDMs~\cite{LDM}, the objective is to estimate noise from latent inputs containing various noise levels. By injecting noise into the latent code, we effectively simulate this noisy distribution. This simulation aligns the latent space more closely with the data distribution encountered during the pre-training phase. Such alignment fosters a more robust and accurate semantic interpretation, which, in turn, enhances the quality of the pseudo labels generated. This shares similar spirits in other applications of diffusion models, such as the text-to-3D~\cite{DreamFusion} where injecting extra noise into data can improve the denoising quality of image and yields better pseudo labels. As shown in Figure~\ref{fig:DPLG}, our DPLG can generate more accurate pseudo-labels compared to the noise-free addition. By strategically incorporating noise in the pseudo-label generation, DPLG enhances the model's adaptability to the target domain and mitigates the bias of semantic understanding. Then, the pseudo-labeled loss is formulated as: 
	\[ \mathcal{L}_{t} = \mathcal{L}_{CE}(p_t, \hat{y}_t) \cdot q. \]
	
	Here, $q$ is the confidence value calculated by the softmax probability of $p_t$~\cite{DAFormer,MIC,DACS}. The consistency regularization of $x_t$ between the teacher and student networks promotes adaptation to the target modality. It encourages the student network to match the teacher's predictions, even under the perturbations introduced by strong data augmentation.
	
	\subsection{Label Palette and Latent Regression}
	TIDMs compress the sample into a latent space with an 8x down-sampling factor, which leads to a serious loss of semantic detail. Specifically, for the input sample with a resolution of $512 \times 512$, it is reduced to $64 \times 64$ after embedded via $\mathcal{E}$. Then, within the denoising UNet decoder, multi-scale features are extracted $f_{s/t}$ after the 5th, 8th, and 11th blocks: $64 \times 64$, $32 \times 32$, and $16 \times 16$. 
	
	For diffusion-based depth estimation~\cite{Marigold}, leveraging the VAE decoder $\mathcal{D}$ to upsample the denoised latent representation back to the original resolution is a natural fit, which recovers the fine-grained scene details. However, the above method is not applicable to semantic segmentation due to the inherent difference between regression and classification. Therefore, to address the problem of semantic detail loss, we propose the Label Palette and Latent Regression (LPLR) module which converts semantic segmentation into regression and utilizes the VAE Decoder $\mathcal{D}$ to obtain high-resolution semantic features.
	
	Initially, the one-hot encoded labels $y_s$ and $\hat{y}_t$ are transformed into a perceptually meaningful RGB space with a pre-defined palette. These RGB representations are then encoded back into the latent space and supervise the UNet's output $o_{s/t} = \mathrm{UNet}(z_{s/t},c)$ in a regression form:
	\[ \mathcal{L}_{s/t}^{reg} = |\mathcal{E}(\mathrm{Palette}(y_s/\hat{y}_t)) - o_{s/t}|. \]
	
	With this supervision, we are able to utilize the VAE decoder $\mathcal{D}$ to obtain a high-resolution semantic regression feature $\mathcal{D}(o_{s/t})$. This feature, combined with the multi-scale features, is then fed to the segmentation head $p_{s/t}=\mathrm{Seg}(f_{s/t}, \mathcal{D}(o_{s/t}))$.
	
	By employing LPLR, we effectively convert the semantic segmentation into a regression problem that can be tackled by the VAE decoder's upsampling capabilities, which retain the fine-grained details necessary for accurate segmentation. Finally, the total training objective is a sum of these losses: 
	\[\mathcal{L} = \mathcal{L}_{s} + \mathcal{L}_{t} + \lambda_{reg}(\mathcal{L}_{s}^{reg} + \mathcal{L}_{t}^{reg}).\]

	\section{Experiments}
	
	\subsection{Implementation Details}
	In our work, we utilize the Stable Diffusion v1-4 model~\cite{LDM}, which has been pre-trained on the LAION-5B~\cite{LAION-5B} dataset, as our TIDM.
	For the segmentation head, we instantiate it with the decoder from DAFormer~\cite{DAFormer}. 
	We train our MADM for 10k iterations with a batch size of 2 and an image resolution of 512$\times$512. 
	The optimization is instantiated with AdamW~\cite{AdamW} with a learning rate of 5e-6.
        For hyperparameters $\beta$, $\gamma$, and $\lambda_{reg}$ in DPLG and LPLR, we set them to \{5000,60,1.0\}/\{8000,50,1.0\}/\{8000,50,10.0\} for depth/infrared/event modalities, respectively.
	In our experiments, we adopt the Cityscapes-Image~\cite{Cityscapes} dataset as the source modality and the DELIVER-Depth~\cite{DELIVER}, FMB-Infrared~\cite{FMB}, and DSEC-Event~\cite{DSEC} datasets as the target modalities.
	Since the semantic classes in these datasets are not identical, we merge some semantically similar classes during training.
	Experiments are conducted on a NVIDIA H800 GPU, occupying about 57G memory.
	
	\begin{table}[t]
		\centering
		\caption{Semantic segmentation results evaluated with MIoU (\%) on three modalities. \textbf{Bold} numbers are the best, \underline{underscored} second best.}
		\label{tab:SoTA}
		
		\begin{subtable}[t]{1.0\linewidth}
			\captionsetup{font={small}} 
			\caption{Cityscapes~\cite{Cityscapes} $\rightarrow$ DELIVER-Depth~\cite{DELIVER}.}
			\centering
			\resizebox{0.9\textwidth}{!}{
				\begin{tabular}{@{}c@{\enspace}|c@{\enspace}c@{\enspace}c@{\enspace}c@{\enspace}c@{\enspace}c@{\enspace}c@{\enspace}c@{\enspace}c@{\enspace}c@{\enspace}c@{\enspace}|c}
					\toprule
                        Method & Sky & Build. & Fence & Person & Pole & Road & S.walk & Veg. & Vehi. & Wall & Tr.S. & MIoU (avg) \\
					\midrule 
					DAFormer~\cite{DAFormer} & 82.28 & 43.35 & 11.82 & \underline{56.03} & 13.90 & \underline{80.10} & 15.44 & 60.08 & 72.67 & \underline{0.18} & \textbf{44.20} & 43.64 \\
					MIC~\cite{MIC}     & 85.10 & 77.78 & 7.30 & 33.41 & \underline{21.14} & 77.04 & 27.24 & \underline{67.07} & 57.25 & 0.00 & \underline{43.92} & 45.21 \\
					PiPa~\cite{PiPa} & 76.90 & \underline{79.65} & \underline{15.61} & \textbf{60.21} & 18.76 & 77.71 & \underline{35.30} & 59.76 & \textbf{84.54} & 0.00 & 31.04 & 49.04 \\
					Rein~\cite{Rein} & \underline{92.00} & 78.78 & \textbf{27.75} & 43.88 & \textbf{32.34} & 78.81 & 27.50 & 58.06 & 76.45 & \textbf{0.34} & 36.68 & \underline{50.23} \\
					MADM & \textbf{95.52} & \textbf{86.70} & 12.48 & 41.88 & 18.99 & \textbf{93.97} & \textbf{54.12} & \textbf{67.12} & \underline{84.29} & 0.00 & 33.34 & \textbf{53.49} \\
					\bottomrule
				\end{tabular}
			}
		\end{subtable}
		
		\begin{subtable}[t]{1.0\linewidth}
			\caption{Cityscapes~\cite{Cityscapes} $\rightarrow$ FMB-Infrared~\cite{FMB}.}
			\centering
			\resizebox{0.9\textwidth}{!}{
				\begin{tabular}{@{}c@{\enspace}|c@{\enspace}c@{\enspace}c@{\enspace}c@{\enspace}c@{\enspace}c@{\enspace}c@{\enspace}c@{\enspace}c@{\enspace}|c}
					\toprule
					Method & Sky & Build. & Person & Pole & Road & S.walk & Veg. & Vehi. & Tr.S. & MIoU (avg) \\
					\midrule
					DAFormer~\cite{DAFormer} & 36.97 & 66.78 & 51.42 & 18.91 & 41.23 & 28.81 & 43.88 & 69.44 & 12.71 & 41.13 \\
					PiPa~\cite{PiPa} & 25.42 & 71.60 & 63.62 & 16.40 & 39.53 & \underline{31.64} & 45.21 & 70.25 & \underline{41.38} & 45.01 \\
					MIC~\cite{MIC}     & 38.11 & \underline{71.63} & 57.89 & 17.59 & 40.68 & \textbf{33.93} & 49.49 & 70.26 & 29.85 & 45.49  \\
					Rein~\cite{Rein} & \underline{84.07} & \textbf{72.84} & \underline{67.10} & \textbf{26.40} & \underline{85.92} & 30.50 & \textbf{72.61} & \textbf{84.51} & 21.95 & \underline{60.65} \\
					MADM & \textbf{88.79} & 71.52 & \textbf{70.51} & \underline{22.30} & \textbf{89.08} & 19.88 & \underline{69.83} & \underline{77.10} & \textbf{51.08} & \textbf{62.23}  \\
					\bottomrule
				\end{tabular}
			}
		\end{subtable}
		
		\begin{subtable}[t]{1.0\linewidth}
			\caption{Cityscapes~\cite{Cityscapes} $\rightarrow$ DSEC-Event~\cite{DSEC}.}
			\centering
			\resizebox{0.9\textwidth}{!}{
				\begin{tabular}{@{}c@{\enspace}|c@{\enspace}c@{\enspace}c@{\enspace}c@{\enspace}c@{\enspace}c@{\enspace}c@{\enspace}c@{\enspace}c@{\enspace}c@{\enspace}c@{\enspace}|c}
					\toprule
					Method & Sky & Build. & Fence & Person & Pole & Road & S.walk & Veg. & Vehi. & Wall & Tr.S. & MIoU (avg) \\
					\midrule
					DAFormer~\cite{DAFormer} & 81.14 & 51.43 & 1.15 & 0.03 & 10.59 & 72.49 & 26.45 & 61.14 & 39.79 & 0.00 & 24.84 & 33.55 \\
					PiPa~\cite{PiPa} & 91.38 & 76.30 & 6.41 & 0.71 & 18.15 & 83.97 & 33.22 & \underline{77.88} & 55.61 & 0.00 & 32.49 & 43.28 \\
					MIC~\cite{MIC}     & \underline{92.36} & \textbf{79.20} & 6.69 & \textbf{32.80} & \underline{19.30} & 79.75 & 31.46 & 68.17 & 58.35 & 0.01 & \textbf{39.30} & 46.13 \\
					Rein~\cite{Rein} & 85.40 & 73.34 & \underline{9.49} & \underline{32.28} & 18.71 & \underline{90.64} & \underline{53.88} & 75.42 & \textbf{79.44} & \underline{12.77} & \underline{39.13} & \underline{51.86} \\
					MADM & \textbf{92.60} & \underline{78.21} & \textbf{26.51} & 29.08 & \textbf{22.78} & \textbf{92.20} & \textbf{62.90} & \textbf{81.70} & \underline{75.11} & \textbf{23.92} & 34.43 & \textbf{56.31} \\
					\bottomrule
				\end{tabular}
			}
		\end{subtable}
	\vspace{-10pt}
	\end{table}

	\subsection{Datasets Setting}

        \textbf{Cityscapes--Image.}
	Cityscapes~\cite{Cityscapes} is the source dataset in our experiments, which constitutes a real-world collection of street-view images captured across 50 distinct urban environments.
	The dataset is split into 2,975 training images and 500 validation images with a resolution of 2048$\times$1024. 
	It provides comprehensive semantic labeling at the pixel-level with 19 distinct semantic classes.
 
        \textbf{DELIVER--Depth.}
	DELIVER~\cite{DELIVER} is a synthetic dataset containing five environmental conditions created by the CARLA simulator~\cite{CARLA}.
	The dataset contains 25 semantic classes and 3,983/2,005/1,897 samples for training/validation/testing with a resolution of 1024$\times$1024.

        \textbf{FMB--Infrared.}
	FMB~\cite{FMB} is an urban street dataset with 1,500 RGB-Infrared pairs at a resolution of 800$\times$600 with 14 semantic classes. 
	It contains a wide range of real driving scenes under different lighting and weather conditions.

        \textbf{DSEC--Event.}
	DSEC~\cite{DSEC} is a stereo event camera dataset for driving scenarios. Driving data are recorded for 3,193 seconds in diverse illumination conditions and urban/rural environments.
	Event data have a resolution of 640$\times$480 with 11 semantic classes and we aggregate them into the edge form in a recurrent manner~\cite{EventsToVideo}.

	\subsection{Comparison with State of the Art Methods}
	Table~\ref{tab:SoTA} presents the comparison with existing SoTA methods DAFormer~\cite{DAFormer}, MIC~\cite{MIC}, PiPa~\cite{PiPa}, and Rein~\cite{Rein} across three modalities: Depth, Infrared, and Event.
	The comparison is based on the Mean Intersection over Union (MIoU) over all classes, a standard measure of segmentation accuracy.
	
	Our MADM demonstrates a strong performance, achieving the MIoU of 53.49\%, 62.23\%, and 56.31\% on the depth, infrared, and event modalities, respectively. 
	It showcases a significant improvement over the SoTA method Rein~\cite{Rein} by +3.26\%, +1.58\%, +4.45\%, which underscores the robustness and effectiveness of MADM in handling the other visual modalities.
	Also, it is worth noting that the self-training loss in our MADM is built upon DAFormer~\cite{DAFormer}, exceeding it average of +17.9\%.

	\begin{figure}[t]
		\centering
		\includegraphics[width=0.8\textwidth]{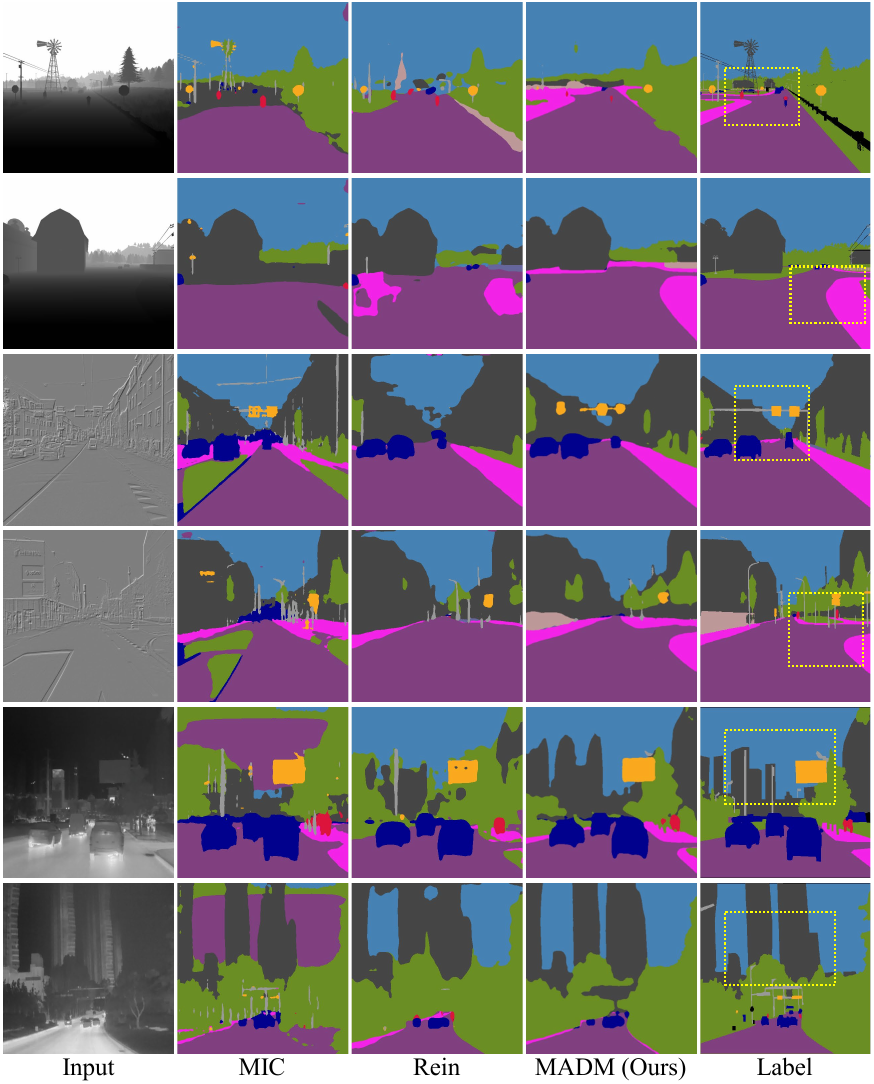}
		\captionsetup{font={small}} 
		\caption{
			Qualitative semantic segmentation results generated by SoTA methods MIC~\cite{MIC}, Rein~\cite{Rein}, and our proposed MADM on three modalities.
		}
		\label{fig:SoTA}
		\vspace{-10pt}
	\end{figure}

	Figure~\ref{fig:SoTA} offers a intuitive comparison of the semantic segmentation results. 
	MIC~\cite{MIC} leverages the SegFormer backbone~\cite{SegFormer} that is pre-trained on the ImageNet-1k dataset~\cite{ImageNet}, enabling it to capture more details within scenes, such as "\textcolor[RGB]{153, 153, 153}{pole}". 
	However, it exhibits weaker modality understanding, leading to frequent mis-segmentation, such as incorrectly classifying the "\textcolor[RGB]{70, 130, 180}{sky}" as the "\textcolor[RGB]{128, 64, 128}{road}".
	In contrast, Rein~\cite{Rein} is built upon the DINOv2 backbone that is pre-trained on extensive, curated datasets without explicit supervision~\cite{DINOv2}. 
	This results in an improved semantic understanding of modalities compared to MIC~\cite{MIC}. 
	Nonetheless, Rein still encounters issues with mis-segmentation and instances of under-segmentation.
	
	Our MADM stands out for its exceptional ability to output precise segmentation results that closely mirror the ground truth. 
	The incorporation of TIDM significantly enhances the generalization of our approach, providing an enhanced comprehension of diverse visual modalities and substantially mitigating the mis-segmentation.

        \begin{wraptable}{r}{0.5\linewidth}
		\centering
		\captionsetup{font={small}} 
		\vspace{-33pt}
		\captionof{table}{Ablation of DPLG and LPLR in depth, infrared, and event modalities.}
		\label{tab:ablation}
		\resizebox{0.99\linewidth}{!}{
			\begin{tabular}{lcccc}
				\toprule
				\multicolumn{1}{l}{Modality} & Baseline & w/ DPLG & w/ LPLR & MADM \\
				\midrule
				Depth & 50.61 & 51.65 & \underline{52.91} & \textbf{53.17}$\pm$0.26 \\ 
				Infrared & 56.28 & \underline{61.86} & 58.75 & \textbf{62.14}$\pm$0.18 \\ 
				Event & 52.27 & 52.84 & \underline{53.05} & \textbf{56.12}$\pm$0.20 \\
				\midrule
				Average & 53.05 & +2.40 & +1.85 & +4.09 \\
				\bottomrule
			\end{tabular}
		}
        \vspace{-10pt}
	\end{wraptable}
	
	\subsection{Ablation Studies}
	Table~\ref{tab:ablation} presents the complete ablation studies that quantify the performance gains achieved by incorporating our proposed DPLG and LPLR into the baseline. 
	The "Baseline" column indicates the performance of the MADM model without DPLG and LPLR.
	It serves as a reference point but achieves performance on par with the SoTA methods in Table~\ref{tab:SoTA}, which demonstrates the strong generalization of TIDMs.
	
	(1) When DGLP is employed, the MIoU is improved by an average of +2.40\%, highlighting the effectiveness of generating robust pseudo-labels.
	Especially in the infrared modality, it achieves a +5.58\% relative improvement over the baseline.
	The quantitative results of the pseudo-labels enhancement are shown in Figure~\ref{fig:DPLG}. 
	(2) The application of LPLR contributes to an average gain of +1.85\%, emphasizing the importance of high-resolution features for segmentation tasks. 
	(3) By employing both DGLP and LPLR, we observe a significant enhancement in +4.09\% over the baseline, which underscores the synergistic benefits of combining robust pseudo-labels generation with fine-grained feature extraction.

	\begin{figure}[t]
		\centering
		\centerline{\includegraphics[width=0.98\textwidth]{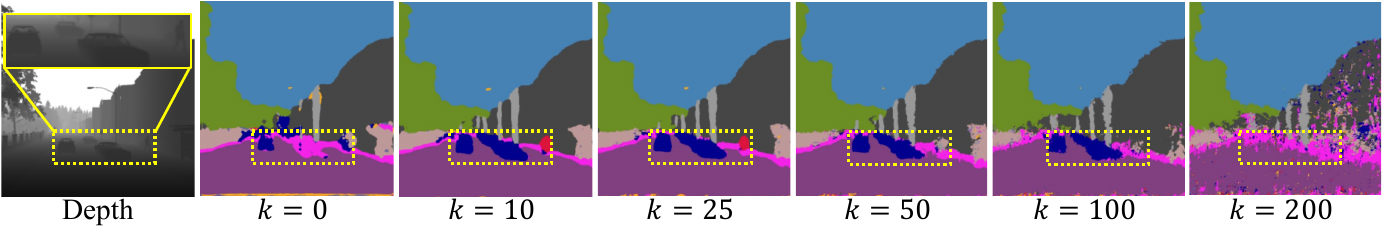}}
		\caption{
			At the 1,250th iteration, we present a visual analysis of diffusion step $k$ in DPLG.
		}
		\label{fig:DPLG_k}
        \vspace{-12pt}
	\end{figure}
	
	\begin{wraptable}{rt}{0.4\linewidth}
		\centering
		\captionsetup{font={small}} 
		\vspace{-12pt}
		\captionof{table}{Ablation of $\beta$ and $\gamma$ in DPLG on the depth modality.}
		\label{tab:DPLG}
		\resizebox{0.99\linewidth}{!}{
			\begin{tabular}{ccccc}
				\toprule
				\multicolumn{1}{c}{\diagbox{$\beta$}{$\gamma$}} & 2,000 & 5,000 & 8,000 & Average \\
				\midrule
				40 & 51.62 & 52.46 & 52.52 & 52.20 \\ 
				60 & 51.30 & \textbf{53.49} & 52.55 & 52.45 \\
				80 & 52.01 & \underline{52.85} & 48.63 & 51.16 \\ 
				\midrule
				Average & 51.64 & 52.93 & 51.23 & - \\
				\bottomrule
			\end{tabular}
		}
        \vspace{-10pt}
	\end{wraptable}
	
	\subsection{Diffusion-based Pseudo-Label Generation}
	We analyze the pivotal roles of $\beta$ and $\gamma$ in our proposed DPLG, particularly within the depth modality. 
	These two parameters control the diffusion step $k$ on $z_t$, which is central to the stability and quality of pseudo-labels.
	
	Table~\ref{tab:DPLG} provides a detailed presentation of the impact of $\beta$ and $\gamma$. 
	For instance, when $\gamma$ is set to 5,000, an increase in $\beta$ from 40 to 60 leads to a noticeable improvement in performance, with the model achieving its peak score of 53.49\%. 
	However, further increasing $\beta$ to 80 results in a decline in performance, indicating the existence of an optimal balance between these parameters.
	
	In Figure~\ref{fig:DPLG_k}, we offer an illustration of how the diffusion step $k$ influences the generation of pseudo-labels. 
	With noise-free addition ($k=0$), the model encounters difficulties in accurately segmenting the "\textcolor[RGB]{0, 0, 142}{car}" and "\textcolor[RGB]{220, 20, 60}{person}" classes. 
	Upon introducing a moderate quantity of noise ($k=10\sim50$), the segmentation is noticeably enhanced, yielding more robust segmentation. 
	Conversely, an excessive amount of noise ($k=200$) leads to a significant degradation in segmentation.

	\subsection{Label Palette and Latent Regression}

	\begin{wraptable}{r}{0.5\linewidth}
		\centering
		\captionsetup{font={small}} 
		\vspace{-15pt}
		\captionof{table}{Ablation of $\lambda_{reg}$ in LPLR on event modality.}
		\label{tab:LPLR}
		\resizebox{0.99\linewidth}{!}{
			\begin{tabular}{cccccc}
				\toprule
				\multicolumn{1}{c}{$\lambda_{reg}$} & 1.0 & 3.0 & 5.0 & 10.0 & 15.0 \\
				\midrule
				MIoU & 53.31 & 54.40 & \underline{55.84} & \textbf{56.31} & 55.31 \\ 
				\bottomrule
			\end{tabular}
		}
        \vspace{-10pt}
	\end{wraptable}
 
	Table~\ref{tab:LPLR} analyzes the loss weight $\lambda_{reg}$ within the proposed LPLR, which regulates the contribution of the regression losses. A minimal $\lambda_{reg}$ of 1.0 and 3.0 yields MIoU of 53.31\% and 54.40\%, respectively, indicating the initial benefits of incorporating regression losses.
	Increasing $\lambda_{reg}$ to 10.0 achieves the optimal MIoU of 56.31\%, signifying the most effective balance between the segmentation and regression losses.
 
	\begin{wrapfigure}{r}{0.49\textwidth}
		\vspace{-13.5pt}
		\centering
		\includegraphics[width=0.47\textwidth]{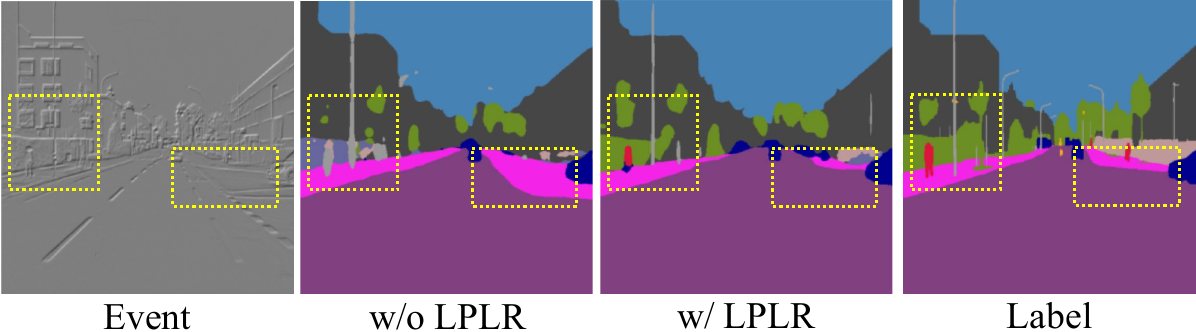}
		\captionsetup{font={scriptsize}} 
		\vspace{-6pt}
		\caption{
			Qualitative Visualization on the event modality w/o and w/ our proposed LPLR. The prediction with LPLR shows more accurate fine-grained segmentation.
		}
		\label{fig:LPLR}
		\vspace{-20pt}
	\end{wrapfigure}
 
	In Figure~\ref{fig:LPLR}, we offer an illustration of the impact of LPLR. It can be seen that the utilization of LPLR results in a more fine-grained segmentation, e.g., "\textcolor[RGB]{220, 20, 60}{person}" and "\textcolor[RGB]{107, 142, 35}{vegetation}" in the left yellow box, "\textcolor[RGB]{128, 64, 128}{road}" and "\textcolor[RGB]{244, 35, 232}{sidewalk}" in the right yellow box, which greatly improves the performance of our MADM.

        \subsection{Benefits of MADM in Nighttime Datasets}
        \label{sec:nighttime}
        
\begin{figure}[h]
            \centering
            \centerline{\includegraphics[scale=0.75]{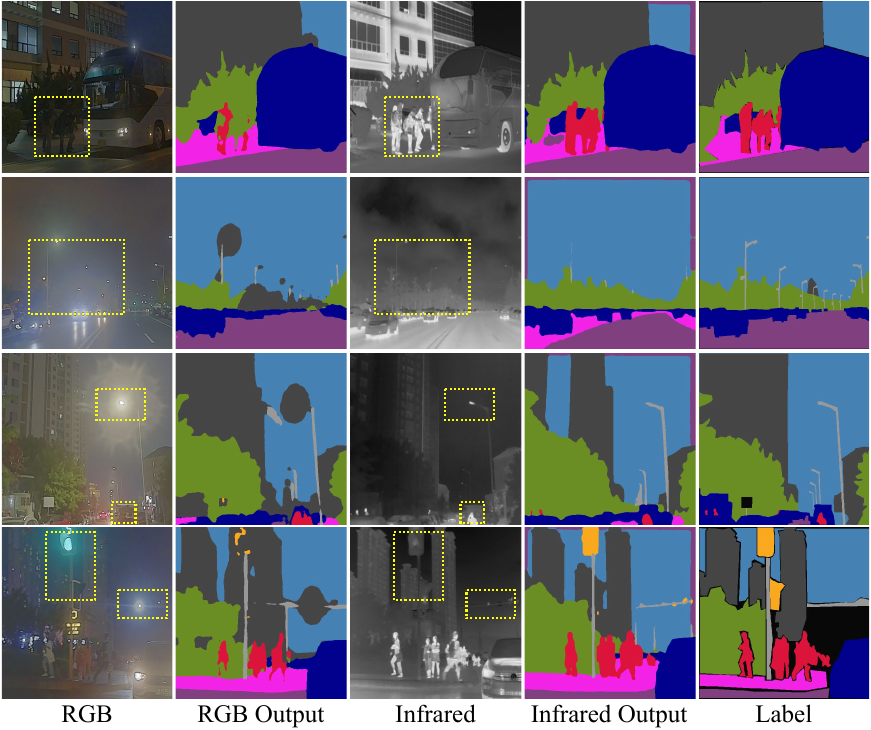}}
            \caption{Visualization of daytime RGB images in Cityscapes dataset~\cite{Cityscapes} $\rightarrow$ nighttime RGB and Infrared modalities in FMB dataset~\cite{FMB}}
            \label{fig:r_night}
        \end{figure}

        As mentioned in Section~\ref{Introduction}, we indicate that other visual modalities present in real-world scenarios are valuable in nighttime perception. In this section, experiments are conducted on the infrared modality to prove this. The FMB-Infrared dataset~\cite{FMB} includes both image and infrared modalities on daytime and nighttime scenes. We adapt from cityscapes~\cite{Cityscapes} with daytime RGB images to the nighttime image modality and infrared modality by our proposed MADA, respectively. Figure~\ref{fig:r_night} and Table~\ref{tab:nighttime} show that the infrared modality has a clear advantage in the "Person" class due to obvious thermal differences and a good suppression of light interference.

        \begin{table}[h]
        \caption{Semantic segmentation results of RGB and infrared modalities evaluated with MIoU (\%) on FMB dataset~\cite{FMB}.}
        \begin{center}
        \begin{tabular}{@{}c@{\enspace}|c@{\enspace}c@{\enspace}c@{\enspace}c@{\enspace}c@{\enspace}c@{\enspace}c@{\enspace}c@{\enspace}c@{\enspace}|c}
        \toprule
        Method & Sky & Build. & Person & Pole & Road & S.walk & Veg. & Vehi. & Tr.S. & MIoU (avg) \\
        \midrule 
        RGB & \textbf{88.85} & 68.14 & 64.79 & \textbf{25.80} & \textbf{89.09} & \textbf{32.43} & 70.32 & \textbf{84.13} & 7.27 & 58.98 \\
        Infrared & 87.94 & \textbf{82.40} & \textbf{82.69} & 21.50 & 76.21 & 26.50 & \textbf{76.61} & 83.80 & \textbf{16.69} & \textbf{61.59} \\
        \bottomrule 
        \end{tabular}
        \end{center}
        \label{tab:nighttime}
        \end{table}

	\section{Conclusion}
	
	In this paper, we present MADM.
	With the powerful generalization of TIDMs, we extend domain adaptation to modality adaptation, aiming to segment other unexplored visual modalities in the real-world.
	Meanwhile, we propose DPLG and LPLR to solve the problems of pseudo-labeling instability and low-resolution features extraction within TIDMs.
	We hope our method can motivate further research on visual modalities other than RGB images.
	\textbf{Limitations:}
	However, despite using only a single-step forward for the diffusion model, the computation far exceeds existing UDASS networks. 
	Future work could focus on distilling the knowledge of TIDMs into lightweight models when adapting.
	\textbf{Broader Impacts:}
	Our work pushes the boundary of semantic segmentation for other visual modalities, which will benefit several applications like multimodal fusion. To the best of our ability, MADM has little to no negative social impact.

        \section*{Acknowledgements}
	This work was supported by National Natural Science Foundation of China (Basic Science Center Program: 61988101), National Natural Science Foundation of China (62233005, 62293502), Fundamental Research Funds for the Central Universities(222202417006), the Programme of Introducing Talents of Discipline to Universities (the 111 Project) under Grant B17017. Pan Zhou was supported by the Singapore Ministry of Education (MOE) Academic Research Fund (AcRF) Tier 1 grants (project ID: 23-SIS-SMU-028 and 23-SIS-SMU-070).
 
	{\small
		\bibliographystyle{unsrt}
		\bibliography{egbib}
	}

	\newpage
	\appendix
	
	\section{Appendix}

        \subsection{Visualization of LPLR}

\begin{figure}[H]
            \centering
            \centerline{\includegraphics[scale=0.6]{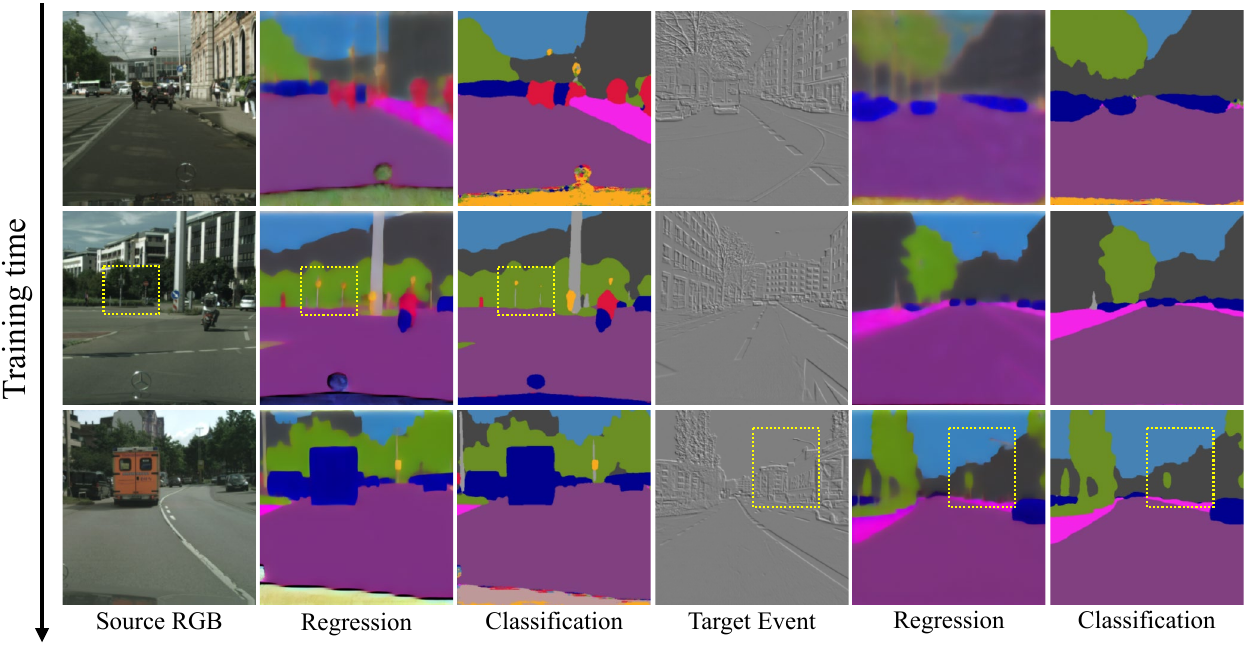}}
            \caption{Visualization of the output of VAE decoder (Regression) and segmentation head (Classification).}
            \label{fig:r_LPLR}
        \end{figure}

        We visualize LPLR under different iteration steps in Figure~\ref{fig:r_LPLR}. "Regression" and "Classification" in Figure~\ref{fig:r_LPLR} denote the output of the VAE decoder and segmentation head, respectively. Our proposed LPLR leverages the up-sampling capability of a pre-trained VAE decoder in a recycling manner. As the model converges, the regression results transform from blurry to progressively clearer states, presenting more details compared to the classification results. This assists the segmentation head in producing more accurate semantic segmentation results. 

        \subsection{Influence of Different Data Volumes}

        \begin{table}[h]
        \caption{Influence on different data volumes testing on the DSEC dataset~\cite{DSEC}.}
        \begin{center}
        \begin{tabular}{cccccc}
        \toprule
        Method & Baseline-100\% & MADM-10\% & MADM-25\% & MADM-50\% & MADM-100\% \\
        \midrule
        MIoU & 52.27 & 53.21 & 53.69 & 54.55 & \textbf{56.31} \\
        \bottomrule
        \end{tabular}
        \end{center}
        \label{tab:data_volumes}
        \end{table}
        
        We train our method with 10\%, 25\%, and 50\% of the total target samples in the event modality. Here, the ``Baseline-100\%'' column indicates the performance of the MADM model without DPLG and LPLR and trained on the whole target samples. The results in Table indicate that our proposed MADM consistently outperforms the baseline across all tested data volumes. Additionally, our MADM is robust and effective even when the dataset size is relatively small. 

        \subsection{Parameters and Costs}

        \begin{table}[h]
        \caption{Comparison of parameters and costs.}
        \label{tab:params}
        \begin{center}
        \begin{tabular}{cccccc}
            \toprule
            \multirow{2}{*}{Method} & Training/Iter. & \multirow{2}{*}{Iteration} & Total training & Params & \multirow{2}{*}{MIoU} \\
                & (seconds) & & (hours) & (million) & \\
            \midrule
            DAFormer~\cite{DAFormer} & 0.36 & 40k & 4.0  & 85  & 33.55 \\
                PiPa~\cite{PiPa} & 1.12 & 60k & 18.7 & 85  & 43.28 \\
            MIC~\cite{MIC} & 0.48 & 40k & 5.3  & 85  & 46.13 \\
            Rein~\cite{Rein} & 1.25 & 40k & 13.9 & 328 & 51.86 \\
            MADM & 1.38 & 10k & 3.8  & 949 & \textbf{56.31} \\
                MADM (Distilled) & 0.46 & 10k & 1.3  & 85  & 54.03 \\
            \bottomrule
        \end{tabular}
        \end{center}
	\end{table}
        
        Table~\ref{tab:params} presents a detailed comparison of training timeper iteration, number of iterations, total training time, parameters, and performance across various methods in the DSEC event modality~\cite{DSEC}, including our MADM and its distilled variant. 

        While MADM does exhibit a higher training time per iteration, the advanced visual prior derived from TIDMs necessitates fewer iteration for adaptation, presenting a minimum total training time. Moreover, MADM achieves a substantial performance improvement, with an MIoU of 57.34\%, surpassing other methods. Recognizing the trade-off in parameter count, we have leveraged our MADM model as a teacher to perform a secondary self-training. This approach has enabled us to distill the knowledge embedded in MADM into a more compact DAFormer model~\cite{DAFormer}, MADM (Distilled), which retains a highMIoU of 54.03\% while significantly reducing parameters to 85M and only increasing the training time by 1.3 hours. Our distilled model demonstrates that it is possible to maintain high performance with reduced computational costs, addressing the concerns raised regarding the parameters and efficiency of MADM.
	
\end{document}